\documentclass[10pt,twocolumn,letterpaper]{article}

\usepackage[pagenumbers]{iccv} 
\usepackage{bm}
\usepackage{svg}
\usepackage{arydshln}
\usepackage[numbers,sort&compress]{natbib}
\definecolor{iccvblue}{rgb}{0.21,0.49,0.74}
\usepackage[breaklinks,colorlinks,allcolors=iccvblue]{hyperref}

\def\paperID{66} 
\def\confName{ICCV}
\def\confYear{2025}

\title{Unsupervised Domain Adaptation via Content Alignment for Hippocampus Segmentation}

\author{
Hoda Kalabizadeh$^{1}$ \quad
Ludovica Griffanti$^{1}$ \quad
Pak-Hei Yeung$^{2}$ \\
Ana I.~L. Namburete$^{1}$ \quad
Nicola K.~Dinsdale$^{1}$ \quad
Konstantinos Kamnitsas$^{1}$\\[4pt]
\small$^{1}$University of Oxford, United Kingdom\\
\small$^{2}$Nanyang Technological University, Singapore\\
{\tt\small hoda.kalabizadeh@cs.ox.ac.uk}
}

\begin{document}
\maketitle
\begin{abstract}
Deep learning models for medical image segmentation often struggle when deployed across different datasets due to domain shifts - variations in both image appearance, known as style, and population-dependent anatomical characteristics, referred to as content. This paper presents a novel unsupervised domain adaptation framework that directly addresses domain shifts encountered in cross-domain hippocampus segmentation from MRI, with specific emphasis on content variations. Our approach combines efficient style harmonisation through $z$-normalisation with a bidirectional deformable image registration (DIR) strategy. The DIR network is jointly trained with segmentation and discriminator networks to guide the registration with respect to a region of interest and generate anatomically plausible transformations that align source images to the target domain.  We validate our approach through comprehensive evaluations on both a synthetic dataset using Morpho-MNIST (for controlled validation of core principles) and three MRI hippocampus datasets representing populations with varying degrees of atrophy. Across all experiments, our method outperforms existing baselines. For hippocampus segmentation, when transferring from young, healthy populations to clinical dementia patients, our framework achieves up to 15\% relative improvement in Dice score compared to standard augmentation methods, with the largest gains observed in scenarios with substantial content shift. These results highlight the efficacy of our approach for accurate hippocampus segmentation across diverse populations.
\end{abstract}    
\section{Introduction}
\label{sec:intro}

Volumetric analysis of brain MRI, particularly of the hippocampus, is a common approach for diagnosing and monitoring dementia ~\cite{mckhann_diagnosis_2011}, with measurements typically obtained through manual or automated segmentation \cite{colliot_discrimination_2008}. Manual segmentation is time-consuming and requires expert knowledge \cite{morey_comparison_2009}, creating a strong demand for accurate automated methods for hippocampus segmentation.

Deep learning (DL) models have demonstrated success in automatic segmentation across various medical imaging modalities \cite{azad_medical_2024}. 
However, their performance generally relies on two key prerequisites: large labelled training datasets and identical distributions between training and testing data. 
In medical applications, these prerequisites are often difficult to meet due to the high cost of expert annotations and distribution mismatch between training (source) and testing (target) domains, known as the \emph{domain shift} \cite{quinonero-candela_dataset_2008}. 
Domain shifts between MRI datasets can be conceptualised as a combination of \emph{style} variations—arising from scanner characteristics and acquisition parameters—and \emph{content} variations, reflecting population differences across studies \cite{guan_domain_2022}. 

\begin{figure}[t]
\centering
    \centering
    \includegraphics[width=\linewidth, clip, trim=3 3 3 3]{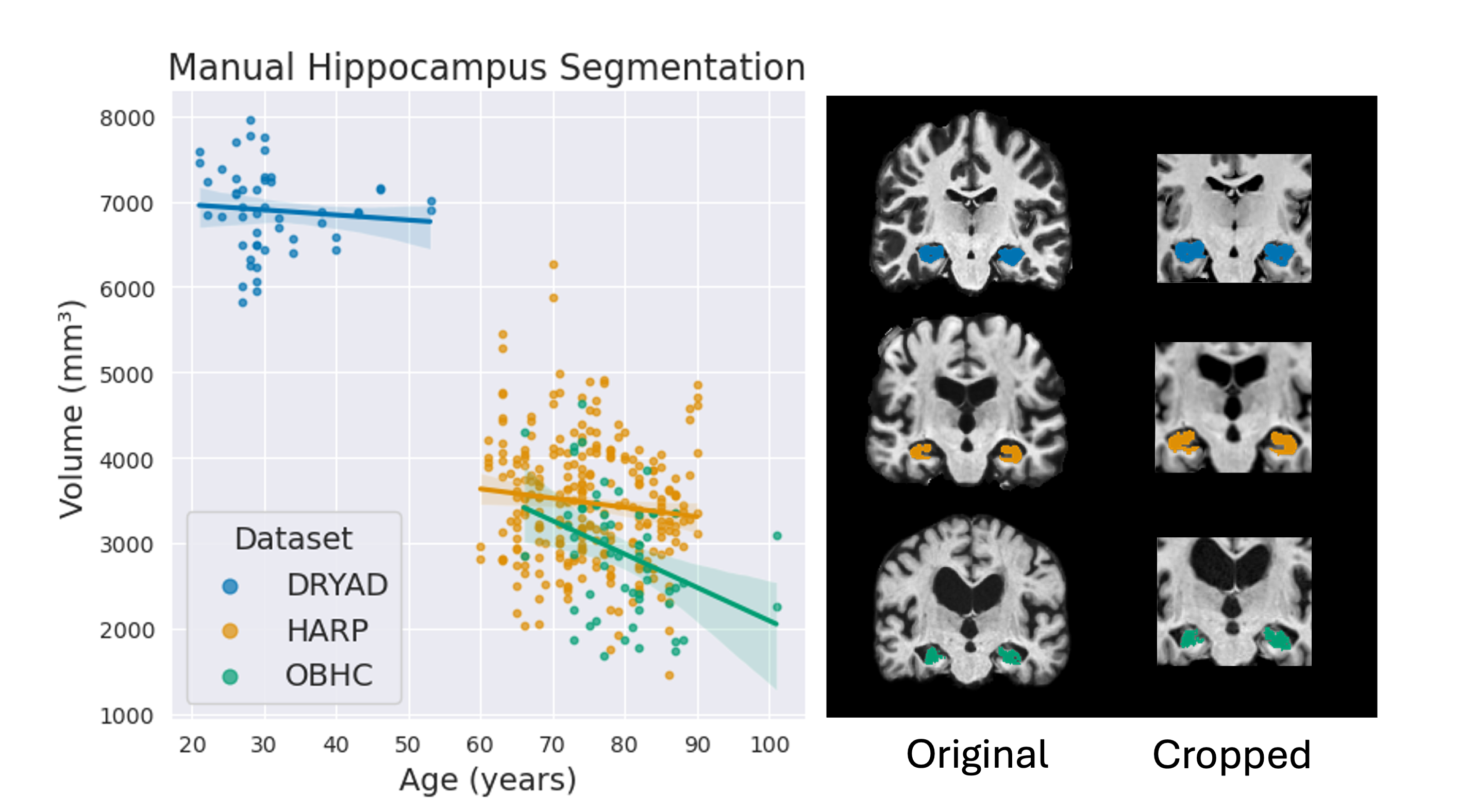}
\caption{Comparison of hippocampal volumes between three populations. DRYAD: young and healthy; HarP: dementia research population; OBHC: dementia clinic population. Demonstrating the shrinking of the hippocampi across populations due to age and disease status/severity.}
\label{hippo_volumes}
\end{figure}

This paper aims to address distribution shifts between related domains in medical imaging, particularly in settings where labels are scarce in the target domain. Specifically, we focus on hippocampus segmentation from MRI across different populations. These datasets exhibit noticeable domain shift in two distinct aspects: \emph{style}, which arises from site and machine differences, and \emph{content}, which is caused by substantial atrophy of the hippocampus and surrounding brain regions due to ageing and disease progression \cite{minkova_gray_2017}. Figure \ref{hippo_volumes} illustrates these variations in hippocampal volumes across datasets with different image acquisitions and population demographics, highlighting the need for domain adaptation methods that address both style and content shifts. 

We aim to tackle the problem through unsupervised domain adaptation (UDA). Existing UDA approaches for medical image segmentation often employ either adversarial feature alignment or generative style transfer methods \cite{kamnitsas_unsupervised_2017,palladino_unsupervised_2020, kumari_deep_2024}. While these techniques effectively address style shifts such as scanner-related variations and imaging protocol differences, they fundamentally assume that the underlying anatomical structures remain consistent across domains. This assumption is violated in cross-population studies where content shifts—including disease progression, ageing, and demographic differences— lead to substantial morphological variations. Moreover, generative approaches often introduce computational overhead and training instability, while feature-level alignment methods may lose important spatial information crucial for dense prediction tasks like segmentation.

To address these limitations, we propose that content shifts should be explicitly modelled in image space through spatial transformations that can capture the underlying anatomical variations between populations. By leveraging deformable registration techniques, we can learn domain-specific anatomical transformations while preserving the spatial correspondence between images and their associated labels—a critical requirement for training segmentation models with transformed data.

To evaluate our approach comprehensively, we conduct experiments on multiple datasets spanning different complexity levels. First, we create a controlled synthetic dataset using a variant of MNIST (Morpho-MNIST \cite{castro_morpho-mnist_2019}) to isolate and validate our method's ability to handle content shifts in a simplified setting, where thick and thin digit variants simulate the structural changes observed in hippocampal atrophy. Second, we demonstrate clinical relevance using three real brain MRI datasets with progressively challenging domain shifts: a young healthy population (DRYAD \cite{kulaga-yoskovitz_multi-contrast_2015}), a dementia research cohort (HarP \cite{boccardi_delphi_2015}), and a clinical dementia population (OBHC \cite{odonoghue_oxford_2023}). As illustrated in Figure \ref{hippo_volumes}, these datasets exhibit substantial hippocampal volume variations across populations, with progressive atrophy from healthy young brains to clinical dementia cases, enabling comprehensive evaluation of our content shift adaptation method across realistic clinical scenarios. \\

In this study, we develop a UDA framework in image space that \emph{explicitly} addresses the domain shifts inherent in cross-domain hippocampus segmentation, with particular attention to content variations. Our contributions include:
\begin{itemize}
    \item We address \emph{style} shift through a robust preprocessing pipeline, eliminating the need for a computationally intensive generative style transfer method.
    \item We propose a DIR-based method for addressing \emph{content} shift, which preserves spatial correspondence between images and their associated labels and can be used to train a downstream segmentation model. 
    \item Extensive experiments across multiple datasets demonstrate noticeable registration and segmentation performance improvements over baselines.
\end{itemize}


\section{Related Work}

\subsection{Unsupervised Domain Adaptation}
Unsupervised domain adaptation (UDA) is a method that aligns source and target domain distributions at either the feature or image level, using only labelled source data and unlabelled target data during training \cite{guan_domain_2022}. 

In feature space, feature alignment methods focus on reducing the disparity between source and target domains by learning domain-invariant features. The goal is to map images from both domains into a shared latent space to minimise discrepancies \cite{kumari_deep_2024}. For medical image segmentation tasks, previous studies have explored implicit discrepancy minimisation approaches using adversarial learning to maximise domain confusion while minimising segmentation loss \cite{kamnitsas_unsupervised_2017, dinsdale_deep_2021}. These adversarial methods have demonstrated significant improvements in multi-site segmentation tasks, effectively addressing \emph{style} shifts that arise from differences in imaging protocols, scanner manufacturers, and acquisition parameters across different medical institutions.

In image space, image translation methods perform domain alignment in pixel space rather than feature space. In previous studies, these methods focused on converting source domain images to match the \emph{style} of the target domain. These approaches commonly employ Generative Adversarial Networks (GANs), particularly CycleGAN \cite{palladino_unsupervised_2020, liu_style_2023}, which enables unpaired image-to-image translation using cycle consistency loss. Other studies have shown that for segmentation tasks, robust preprocessing methods, such as $z$-normalisation, perform comparably to GAN-based style transfer approaches \cite{kalabizadeh_is_2025}. 

Moreover, despite most studies introducing domain shifts occurring due to style and content shifts, many assumed the content/anatomical information to be domain-invariant. This assumption also forms the underlying basis for studies investigating style-content disentanglement \cite{huang_multimodal_2018, lee_drit_2020, dewey_disentangled_2020}, which posit that style is domain-specific while content remains domain-invariant. 

Consequently, UDA methods that explicitly address content shift by treating it as domain-specific have remained largely unexplored. To fill this gap, our framework proposes to model content shifts explicitly in image space through learnable transformations rather than treating them as invariant features, while handling style shifts through preprocessing, thereby avoiding the computational overhead of generative approaches.

\subsection{Deformable Image Registration}

Deformable Image Registration (DIR) is a technique for estimating non-rigid voxel correspondences between fixed and moving image pairs. It relies on two core loss functions: a dissimilarity function between the deformed and fixed images, and the regularisation penalty of the registration field. To preserve topology, additional constraints are commonly imposed on the registration field to ensure the mapping is differentiable and invertible, referred to as diffeomorphic registration \cite{avants_symmetric_2008}. Given the advancements of DL methods, neural network architectures have been developed that take fixed and moving images as input to estimate a registration field \cite{balakrishnan_unsupervised_2018,balakrishnan_voxelmorph_2019,kim_cyclemorph_2021}. A popular example of an unsupervised DL-based registration model is VoxelMorph \cite{dalca_unsupervised_2018,balakrishnan_voxelmorph_2019}. Often proposed for 3D MR brain scans, VoxelMorph learns a registration function for aligning input image pairs.

Inspired by DIR, new methods have emerged for data augmentation via learning deformable transformations. Deformation-based transformations are able to capture more general spatial variations, making them an ideal approach for directly modelling the content shift between source and target data. Recently, studies have investigated learning shape-based data augmentations via deformation-based transformations for medical image segmentation tasks \cite{zhao_data_2019,shen_anatomical_2020}. For example, Zhao et al., \cite{zhao_data_2019} proposed a model for independently learning the spatial transformations between a pair of labelled and unlabelled MR datasets. They then used the model to generate augmented data based on the underlying spatial variations, which was subsequently used to train a segmentation model. Building on these insights, our framework leverages DIR techniques to explicitly model content shifts through learnable spatial transformations, while ensuring that the spatial correspondence between images and labels is preserved—a critical requirement for effective cross-domain segmentation in the presence of anatomical variations.


\subsection{Joint Registration and Segmentation}
Registration and segmentation tasks have great complementarity, which can bring mutual improvement in complex scenes with limited annotations \cite{he_deep_2020-1}. The registration model generates augmented training data through image warping (creating warped image-label pairs) to enhance segmentation training \cite{zhao_data_2019}. This approach reduces annotation requirements while improving segmentation generalisation. Conversely, the segmentation model provides region-based constraints \cite{xu_deepatlas_2019} that focus registration attention on regions of interest, enabling more precise alignment in complex anatomical structures. 

Previous studies have utilised DIR-based learning for one-shot or few-shot segmentation \cite{zhao_data_2019,wang_lt-net_2020}. However, their approach is not optimised for the segmentation task, allowing background structures, unrelated to the task, to have a substantial influence on the overall alignment. To address this limitation, He et al. proposed a complementary joint model for registration and segmentation, suggesting that region constraints fed back via the segmentation model can provide attention on ROIs, resulting in finer registration in complex settings \cite{he_deep_2020-1}. Moreover, they include a discriminator network for adversarial training to further guide alignment by evaluating the similarity between transformed and target images. In another study, Wang et al. propose learning reversible voxel-wise correspondence for one-shot segmentation, investigating bidirectional warping to make the learning process more robust \cite{wang_lt-net_2020}.

While these studies have made significant contributions to one-shot and few-shot segmentation, they do not inherently address the problem of domain shift, as they focus on learning from limited data within the same domain. Our framework bridges this gap by extending the proven complementarity of registration and segmentation to cross-domain scenarios. Here, the registration component learns population-specific anatomical transformations that enable effective domain adaptation rather than merely addressing data scarcity, while segmentation provides anatomically-informed constraints to enable systematic handling of cross-population morphological variations.
\section{Method}
\label{sec:formatting}

\begin{figure*}[t]
  \centering
  \includegraphics[width=0.75\linewidth]{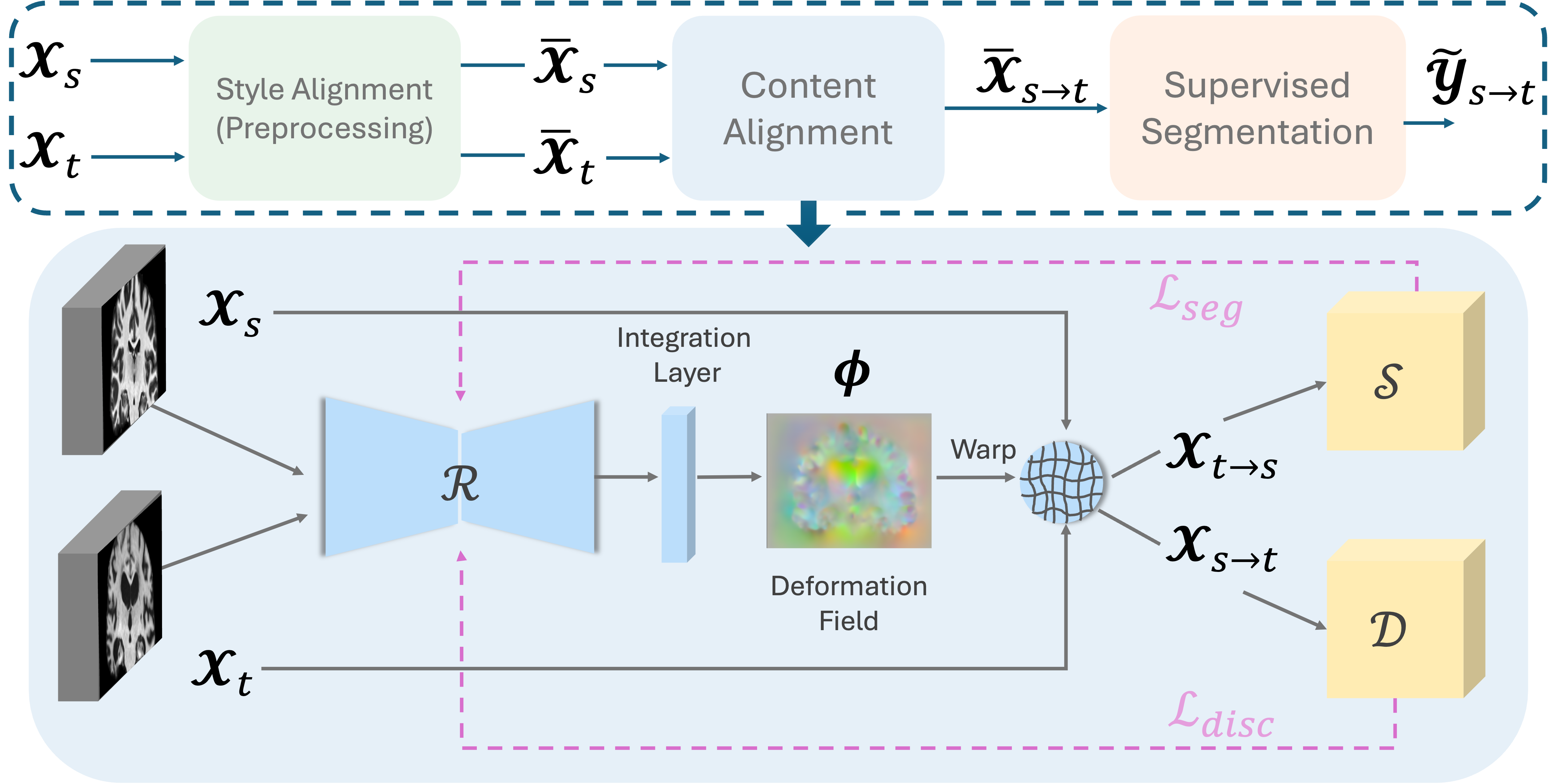}  
  \caption{Top: Overview of the proposed framework, showing the sequence of steps for input source ($\bm{\mathcal{X}}_s$) and target ($\bm{\mathcal{X}}_t$) images to output style and content aligned images that are used to train a downstream segmentation network. Bottom: Proposed model architecture for learning content alignment, formed of registration ($\mathcal{R}$), segmentation ($\mathcal{S}$) and discriminator ($\mathcal{D}$) networks for joint training. }
    \label{model_arch}
\end{figure*}


\subsection{Overview}
Figure \ref{model_arch} visualises the proposed framework. We assume access to a source dataset, $\mathcal{D}_s = \{\bm{\mathcal{X}}_s, \bm{\mathcal{Y}}_s\}$, and an unlabelled target dataset, $\mathcal{D}_t = \{\bm{\mathcal{X}}_t\}$. The framework consists of three core components: preprocessing for style alignment, content alignment, and downstream supervised segmentation. Source and target images are initially preprocessed to address the style shift between domains. 
Following this, the style-aligned images are used to train the content alignment module, which outputs source images aligned to the target domain, with attention focused on the region of interest (ROI) to be segmented, aiding the training process. Lastly, the transformed images and corresponding labels are used to train a segmentation network, which is then evaluated on unseen target images.

\subsection{Data Preprocessing and Style Alignment}
All MR scans were affine registered to a standard space (MNI) to ensure spatial correspondence across participants. The scans also underwent brain extraction to isolate brain tissue from surrounding non-brain structures. N4 bias field correction was then applied to address low-frequency intensity non-uniformities. 

Inspired by findings from \cite{kalabizadeh_is_2025} demonstrating that robust intensity normalisation techniques sufficiently address style variations without requiring the computational complexity of GANs, we applied $z$-score normalisation along with histogram standardisation to harmonise intensity distributions, thereby reducing the impact of style differences. While preprocessing steps mitigate style-related domain shifts, they do not address the more challenging content shift. To overcome this, we developed a DIR-based model detailed in the following section.


\subsection{Deformable Image Registration for Content Alignment}

\subsubsection{Registration Network:}
To train a DIR network,  $\mathcal{R}$, that learns to perform non-linear transformations from source to target images, we utilise the diffeomorphic VoxelMorph \cite{dalca_unsupervised_2019} backbone. $\mathcal{R}$ takes as input the source and target images, $\{\bm{\mathcal{X}}_s, \bm{\mathcal{X}}_t\}$, and outputs a velocity field which is passed through an integration layer to output a diffeomorphic deformation field, $\bm{{\mathcal{\phi}}}$. We implement bidirectional warping by also computing the inverse deformation field $\bm{{\mathcal{\phi}^{-1}}}$. Subsequently,  $\{\bm{\mathcal{X}}_s,\bm{\mathcal{Y}}_s, \bm{\mathcal{X}}_t\}$ are passed into a spatial transform model, with the deformation fields to compute: 
$ \bm{{\mathcal{X}}}_{s \rightarrow t} = \bm{\mathcal{X}}_{s}  \circ \phi$; 
$ \bm{{\mathcal{Y}}}_{s \rightarrow t} = \bm{\mathcal{Y}}_{s}  \circ \phi$, and
$ \bm{{\mathcal{X}}}_{t \rightarrow s} = \bm{\mathcal{X}}_{t}  \circ \phi^{-1}$.
The registration network is optimised using two key loss components: $\mathcal{L}_{sim}$, a similarity loss that uses mean squared error (MSE) to ensure the deformed images closely match their targets in both forward and inverse directions; and $\mathcal{L}_{smooth}$, a smoothness constraint implementing $L2$-regularisation on both the forward and inverse deformation fields to prevent physiologically implausible transformations and ensure spatial consistency of the registration.
\begin{align}
\mathcal{L}_{sim} &= \mathcal{L}_{mse}(\bm{\mathcal{X}}_{s \rightarrow t}, \bm{\mathcal{X}}_{t}) 
+ \mathcal{L}_{mse}(\bm{\mathcal{X}}_{t \rightarrow s}, \bm{\mathcal{X}}_{s}) \\
\mathcal{L}_{smooth} &= \mathcal{L}_{L2}(\bm{\phi}) + \mathcal{L}_{L2}(\bm{\phi}^{-1})
\label{eq:loss_sim}
\end{align}

\subsubsection{Joint Learning Strategy:}
We incorporated a discriminator network, $\mathcal{D}$, that was trained concurrently with the registration network in an adversarial framework. The discriminator learned to differentiate between the deformed source images $({\bm{{\mathcal{X}}}_{s \rightarrow t}})$ and real target images $(\bm{\mathcal{X}}_{t})$ while simultaneously providing adversarial loss feedback to guide the registration process. This feedback was implemented as a binary cross-entropy loss:
$\mathcal{L}_{disc} = \mathcal{L}_{bce}(\bm{{\mathcal{X}}}_{s \rightarrow t})$. The adversarial training encourages the registration network to generate transformations that produce more target-like appearances, overcoming the limitations of conventional similarity metrics by providing additional constraints for content alignment. 

We also included a 3D U-Net segmentation network, $\mathcal{S}$, which primarily provides feedback to the registration network while undergoing training concurrently on the labelled source data. For each training iteration, $\mathcal{S}$ processes two images. First, it takes the source image ($\bm{\mathcal{X}}_s$) as input and outputs the predicted segmentation mask ($\tilde {\bm{\mathcal{Y}}_s}$). A dice loss, $\mathcal{L}_{dice}(\bm{\tilde{\mathcal{Y}}}_{s}, \bm{{\mathcal{Y}}}_{s})$, is computed between the predicted and ground truth mask, which is used to train the segmenter $\mathcal{S}$. Subsequently, the segmenter processes the inverse-deformed target image ($\bm{{\mathcal{X}}}_{t \rightarrow s}$), output by the registration network, predicting the segmentation mask ($\bm{\tilde {\mathcal{Y}}}_{t \rightarrow s}$). This approach allows utilisation of the ground truth source labels, resulting in a Dice loss that is backpropagated to the registration network: $\mathcal{L}_{seg} =  \mathcal{L}_{dice}(\bm{\tilde {\mathcal{Y}}}_{t \rightarrow s}, \bm{{\mathcal{Y}}}_{s})$.

\subsubsection{Combined Loss Function:}
The network is trained using a combined loss, formed using a similarity loss $ \mathcal{L}_{sim}$, smoothing loss $\mathcal{L}_{smooth}$, discriminator loss $\mathcal{L}_{disc}$ and segmentation loss $\mathcal{L}_{seg}$, weighted by $\lambda_{sim}$, $\lambda_{smooth}$, $\lambda_{disc}$ and $\lambda_{seg}$ respectively, resulting in the following objective function:

\begin{align}
\mathcal{L}(\mathcal{R,S,D}) &= \lambda_{sim} \mathcal{L}_{sim} + \lambda_{smooth} \mathcal{L}_{smooth} \nonumber \\
&\quad + \lambda_{disc} \mathcal{L}_{disc} + \lambda_{seg} \mathcal{L}_{seg}
\label{eq:loss_ST}
\end{align}

Once $\mathcal{R}$ was trained, we used the transformed images to train a segmentation model $\mathcal{F}(\bm{{\mathcal{X}}}_{s \rightarrow t},\bm{{\mathcal{Y}}}_{s \rightarrow t})$, such that the performance for $\mathcal{D}_t$ was maximised.

\begin{figure}[b]
\centering
\includegraphics[width=\linewidth]{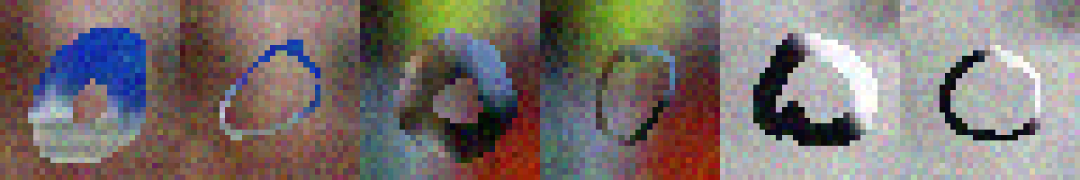}
\caption{Example of simulated Morpho-MNIST thick and thin pairs.}
\label{fig:mnist_example}
\end{figure}

\section{Experimental Setup}
\begin{table*}[t]
\centering
\scriptsize
\setlength{\tabcolsep}{4pt} 
\renewcommand{\arraystretch}{1.2} 
\begin{tabular}{l c c c}
\multicolumn{1}{l}{\textbf{Method}} & \textbf{Training Data} & \textbf{Morpho-MNIST Thick} & \textbf{Morpho-MNIST Thin} \\ \hline
U-Net & Target & 0.981 ± 0.001 & 0.938 ± 0.001  \\ \hline
U-Net & Src (MNIST-Thick) & - & 0.669 ± 0.037  \\ \hline
Image-Space Aug & Src (MNIST-Thick) + Intensity Aug & - & 0.606 ± 0.014  \\ 
Image-Space Aug  & Src (MNIST-Thick) + Affine Aug & - & 0.528 ± 0.025   \\ 
Image-Space Aug  & Src (MNIST-Thick) + Deformable Aug & - & 0.647± 0.020  \\ \hline
Feature-Space UDA \cite{dinsdale_deep_2021} & Src (DRYAD) + Target & - & 0.648 ± 0.163 \\ \hline
\end{tabular}
\caption{DSC for segmentation performance using common augmentation methods in image space, and a representative UDA method for segmentation in feature space. Tested on Morpho-MNIST Thick and Morpho-MNIST Thin.}
\label{Table:Harp_BHC_domain_shift}
\label{domain_shift_MNIST}
\end{table*}

\begin{table*}[t]
\centering
\scriptsize
\setlength{\tabcolsep}{4pt} 
\renewcommand{\arraystretch}{1.2} 
\begin{tabular}{l c c c c}
\multicolumn{1}{l}{\textbf{Method}} & \textbf{Training Data} & \textbf{DRYAD} & \textbf{HarP} & \textbf{OBHC} \\ \hline
U-Net & Target & 0.880 ± 0.012 & 0.854 ± 0.025 & 0.811 ± 0.027 \\ \hline
U-Net & Src (DRYAD) & - &  0.705 ± 0.135 &  0.550 ± 0.153 \\ \hline
Image-Space Aug & Src (DRYAD) + Intensity Aug & - & 0.719 ± 0.083 &   0.552 ± 0.172 \\ 
Image-Space Aug & Src (DRYAD) + Affine Aug & - & 0.717 ± 0.113 & 0.592 ± 0.150 \\ 
Image-Space Aug & Src (DRYAD) + Deformable Aug & - & 0.710 ± 0.126 & 0.562 ± 0.217 \\  \hline
Feature-Space UDA \cite{dinsdale_deep_2021} & Src (DRYAD) + Target & - & 0.686 ± 0.123 & 0.477 ± 0.170\\ \hline
\end{tabular}
\caption{DSC for segmentation performance using common augmentation methods in image space, and a representative UDA method for segmentation in feature space. Models tested on DRYAD, HarP and OBHC.} \label{Table:Harp_BHC_domain_shift}
\label{domain_shift_MRI}
\end{table*}
\subsection{Datasets}
\textbf{Morpho-MNIST Dataset:} We created a synthetic content shift scenario using the Morpho-MNIST dataset \cite{castro_morpho-mnist_2019} to develop and validate our approach before applying it to complex brain MRI data. We obtained thick (source) and thin (target) images of digit zero, where the thinning creates a \emph{content} shift that simulates hippocampal atrophy patterns observed in dementia progression. Additionally, to introduce \emph{style} shift mimicking the variability in real brain scans, we added random image patches from CIFAR \cite{krizhevsky_learning_2009} as background (patch size=5×5) and foreground (patch size=15×15), along with Gaussian noise (mean=0, std=0.05). This approach enables rigorous method validation before transitioning to the more computationally intensive brain MRI applications. Examples of simulated thick and thin image pairs are shown in Figure \ref{fig:mnist_example}. \\




\begin{figure*}[t]
\centering
\includegraphics[width=0.9\textwidth]{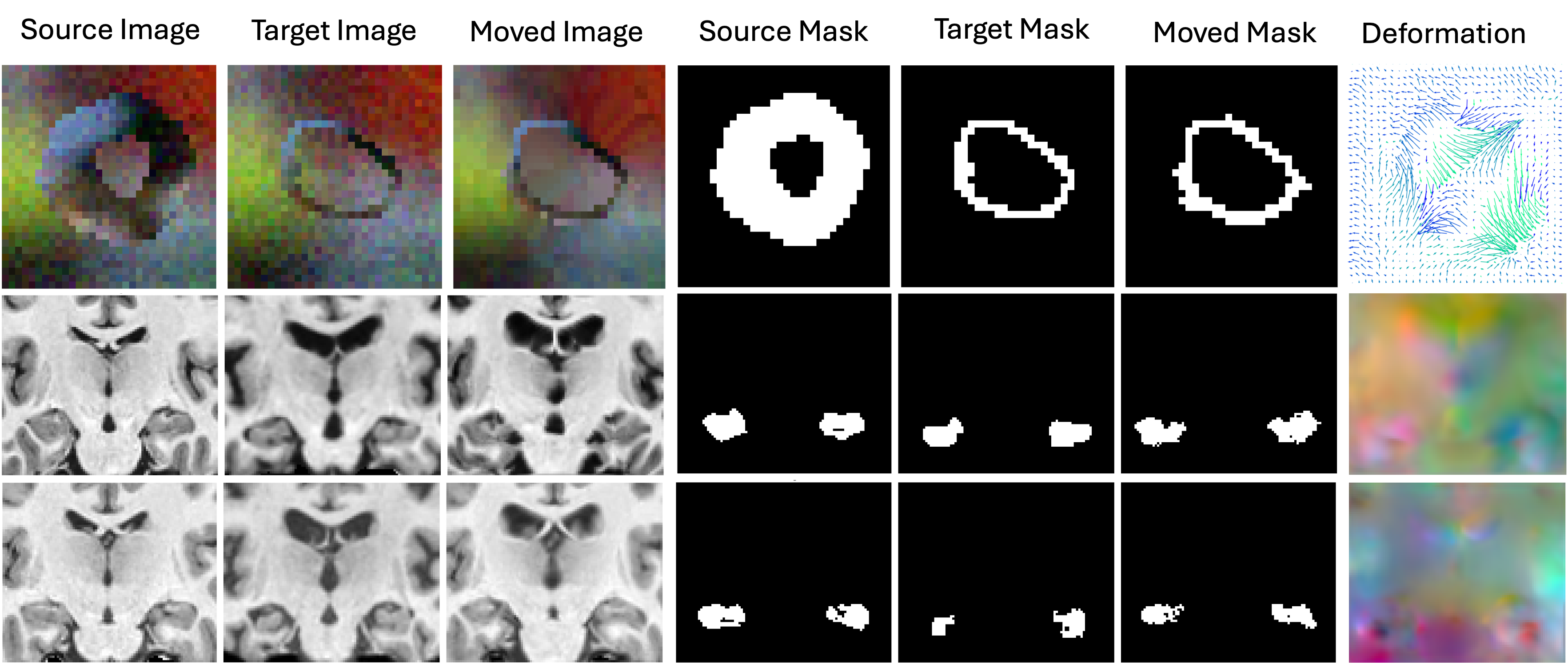}
\caption{Examples of content alignment for Morpho-MNIST thick$\rightarrow$thin (top row), DRYAD$\rightarrow$HarP (middle row), DRYAD$\rightarrow$OBHC (bottom row).}
\label{fig:reg_all}
\end{figure*}
\textbf{Hippocampus Dataset:} To demonstrate the content shift that occurs due to hippocampus atrophy between healthy and dementia populations, we focused on three datasets: Young-Healthy (DRYADS \cite{kulaga-yoskovitz_multi-contrast_2015}), Dementia Research (HarP \cite{boccardi_delphi_2015}) and Dementia Clinical (OBHC \cite{odonoghue_oxford_2023}). These datasets present both style and content shifts. Specifically, DRYAD $\rightarrow$ HarP represents a transition from young healthy brains to an older population with mixed cognitive status, including both healthy individuals and those with dementia. DRYAD $\rightarrow$ OBHC represents a more extreme clinical scenario, transitioning directly from young healthy participants to elderly patients in a real-world clinical setting. Our experiments with these domain pairs allow us to evaluate our method under increasingly challenging and clinically relevant conditions. The three datasets are characterised as follows:

\begin{itemize}
    \item \textbf{Young Healthy Dataset (DRYADS):} 25 young, healthy participants scanned using a 3T Siemens scanner with manually annotated hippocampus labels \cite{kulaga-yoskovitz_multi-contrast_2015}.
    
    \item \textbf{Dementia Research Dataset (HarP):} 130 T1-weighted MRI volumes with manually labelled hippocampi, including cognitively healthy controls, mild cognitive impairment (MCI), and Alzheimer's disease (AD) patients acquired using various scanners \cite{boccardi_delphi_2015, boccardi_training_2015}.
    
    \item \textbf{Dementia Clinical Dataset (OBHC):} 29 consented memory clinic patients from the Oxford Brain Health Clinic, representing real-world clinical diversity. Images acquired on a 3T Siemens scanner with hippocampi manually annotated by an experienced researcher \cite{odonoghue_oxford_2023}.
\end{itemize}
Figure \ref{hippo_volumes} illustrates the substantial anatomical variations across our three datasets and the progressive changes between populations. The left panel presents a scatter plot of age versus manually annotated hippocampal volumes, clearly demonstrating the demographic and volumetric differences between datasets. Notably, there is a pronounced decline in hippocampal volume from the young-healthy cohort (DRYAD) to the dementia populations (HarP and OBHC), reflecting the combined effects of ageing and disease progression. The right panel displays representative brain scans with corresponding hippocampal segmentation masks from each population, visualising the characteristic neuroanatomical changes: enlarged ventricular spaces and reduced hippocampal volumes that accompany advancing age and dementia severity.
\subsection{Implementation Details} 
To speed up training time, the brains were cropped to a size of (96, 128, 80), as visualised in Figure \ref{hippo_volumes}, removing as much of the background and outer areas of the brain as possible. The content alignment and downstream segmentation tasks were evaluated using the following evaluation metrics: Dice score (DSC), Surface Dice score (SDSC), Hausdorff distance (HD), and Relative Absolute Volume Difference (RAVD). For the downstream segmentation task, images were split into left and right hemispheres. We used the Adam optimiser with a learning rate of $1\times 10^{-4}$, $3\times 10^{-4}$ and $1\times 10^{-3}$ for training $\mathcal{R}$, $\mathcal{D}$ and $\mathcal{S}$, respectively. For Equation \ref{eq:loss_ST}, we performed ablation studies on both the Morpho-MNIST and MRI datasets and obtained the best performance with the following loss weights:\\
\textbf{Morpho-MNIST:} $\lambda_{sim} = 1$, $\lambda_{smooth} = 0.001$, $\lambda_{disc} = 0.0001$, $\lambda_{seg} = 0.01$\\
\textbf{MRI:} $\lambda_{sim} = 0.1$, $\lambda_{smooth} = 1$, $\lambda_{disc} = 0.0001$, $\lambda_{seg} = 1$\\
Training was conducted with a batch size of 1 using 3-fold cross-validation. On an Nvidia A10 GPU, content alignment took approximately 40 minutes on Morpho-MNIST and 1.5 hours on MRI, while downstream segmentation required 30 minutes and 1 hour, respectively.

\section{Results \& Discussion}

\begin{table*}[t]
    \centering
    \scriptsize
    \setlength{\tabcolsep}{3.5pt}
    \renewcommand{\arraystretch}{1.2} 
    
    \begin{tabular}{llcccc}
        \textbf{Task} & \textbf{Model} & \textbf{Avg DSC ↑} & \textbf{Avg SDSC ↑} & \textbf{95 \% HD ↓} & \textbf{RAVD↓} \\ \hline
        CA & DIR & 0.758±0.001 & 0.910±0.000 & 2.037±0.012 & 58.290±0.101 \\ 
        CA & + $\mathcal{L}_{disc}$ & 0.777±0.004 & 0.923±0.002 & 1.952±0.031 & 44.749±1.083 \\
        CA & + $\mathcal{L}_{seg}$ & 0.791±0.003 & 0.928±0.003 & 1.742±0.023 & 49.943±1.401  \\
        CA & + $\mathcal{L}_{disc}$ + $\mathcal{L}_{seg}$ & 0.818±0.002 & 0.947±0.003 & 1.540±0.029 & 35.153±2.018 \\ \hline
        DS & Trg (in-domain) & 0.938±0.001 & 0.988±0.001 & 0.632±0.018 & 5.456±0.268 \\
        DS & Src (out-of-domain) & 0.669±0.037 & 0.909±0.007 & 2.109±0.137 & 64.833±11.277 \\ \hdashline
        DS & DIR & 0.725±0.053 & 0.843±0.054 & 3.161±1.256 & 100.279±33.005 \\
        DS & + $\mathcal{L}_{disc}$ & 0.802±0.036 & 0.908±0.028 & 1.846±0.313 & 59.118±17.084 \\
        DS & + $\mathcal{L}_{seg}$ & 0.797±0.044 & 0.909±0.038 & 2.217±0.734 & 58.571±21.941 \\
        DS & + $\mathcal{L}_{disc}$ + $\mathcal{L}_{seg}$ & \textbf{0.830±0.026} & \textbf{0.936±0.018} & \textbf{1.640±0.285} & \textbf{43.588±12.746} \\ \hline
    \end{tabular}
    \caption{Content alignment (CA) and downstream segmentation (DS) performance for Morpho-MNIST thick $\rightarrow$ thin, tested on thin. Bold: best performance.}
    \label{table:MNIST}
\end{table*}

\begin{table*}[t]
    \centering
    \scriptsize
    \setlength{\tabcolsep}{3.5pt}
    \renewcommand{\arraystretch}{1.2} 
    \begin{tabular}{llcccc}
        \textbf{Task} & \textbf{Model} & \textbf{Avg DSC ↑} & \textbf{Avg SDSC ↑} & \textbf{95 \% HD ↓} & \textbf{RAVD↓} \\ \hline
        CA & DIR & 0.566±0.013 & 0.646±0.013 & 4.642±0.140 & 52.007±1.009 \\ 
        CA & + $\mathcal{L}_{disc}$ & 0.573±0.014 & 0.654±0.014 & 4.615±0.140 & 51.900±1.007 \\
        CA & + $\mathcal{L}_{seg}$ & 0.623±0.013 & 0.742±0.012 & 3.702±0.157 & 29.248±1.862 \\
        CA & + $\mathcal{L}_{disc}$ + $\mathcal{L}_{seg}$ & 0.619±0.008 & 0.745±0.014 & 3.598±0.154 & 24.188±3.836\\ \hline
        DS & Trg (in-domain) & 0.854±0.025 & 0.959±0.020 & 1.298±0.425 & 6.468±4.635 \\
        DS & Src (Out-of-Domain) & 0.705±0.135 & 0.803±0.136 & 8.740±19.892 & 19.943±18.797 \\ \hdashline
        DS & DIR & 0.740±0.166 & 0.842±0.160 & 3.492±6.605 & 17.194±20.587 \\
        DS & + $\mathcal{L}_{disc}$ & 0.758±0.115 & 0.864±0.087 & 4.489±12.188 & 14.591±14.402 \\
        DS & + $\mathcal{L}_{seg}$ & 0.764±0.142 & 0.875±0.110 & 3.446±8.376 & 12.863±16.999 \\        
        DS & + $\mathcal{L}_{disc}$ + $\mathcal{L}_{seg}$ & \textbf{0.771±0.096} & \textbf{0.880±0.071} & \textbf{2.696±6.732} & \textbf{10.967±12.764} \\ \hline
    \end{tabular}
    \caption{Content alignment (CA) and downstream segmentation (DS) performance for DRYAD $\rightarrow$ HarP, tested on HarP. Bold: best performance.}
    \label{table:DRYAD-HARP}
\end{table*}

\begin{table*}[t]
    \centering
    \scriptsize
    \setlength{\tabcolsep}{3.5pt}
    \renewcommand{\arraystretch}{1.2} 
    \begin{tabular}{llcccc}
        \textbf{Task} & \textbf{Model} & \textbf{Avg DSC ↑} & \textbf{Avg SDSC ↑} & \textbf{95 \% HD ↓} & \textbf{RAVD↓} \\ \hline
        CA & DIR & 0.514±0.003 & 0.566±0.003 & 6.408±0.02	& 89.627±1.337 \\ 
        CA & + $\mathcal{L}_{disc}$ & 0.510±0.006 & 0.562±0.006 & 6.395±0.043 & 88.353±0.157 \\
        CA & + $\mathcal{L}_{seg}$  & 0.615±0.006 & 0.727±0.005 & 4.822±0.161 & 32.546±3.299 \\
        CA & + $\mathcal{L}_{disc}$ + $\mathcal{L}_{seg}$ & 0.575±0.048 & 0.673±0.073 & 5.185±0.716 & 48.283±32.003 \\ \hline
        DS & Trg (in-domain) & 0.811±0.027 & 0.917±0.028 & 4.462±0.550 & 10.714±7.566 \\
        DS & Src (Out-of-Domain) & 0.550±0.153 & 0.661±0.133 & 8.929±4.211 & 37.311±24.647 \\ \hdashline
        DS & DIR & 0.616±0.233 & 0.739±0.221 & \textbf{7.018±3.894} & 27.716±30.375 \\
        DS & + $\mathcal{L}_{disc}$ & 0.678±0.125 & 0.784±0.104 & 7.419±14.193 & 17.812±15.125 \\
        DS & + $\mathcal{L}_{seg}$ & 0.674±0.078 & 0.777±0.072 & 9.22±23.128 & 16.734±14.750 \\
        DS & + $\mathcal{L}_{disc}$ + $\mathcal{L}_{seg}$ & \textbf{0.680±0.095} & \textbf{0.786±0.082} & 11.182±9.219 & \textbf{16.062±13.607} \\ \hline
    \end{tabular}
    \caption{Content alignment (CA) and downstream segmentation (DS) performance for DRYAD$\rightarrow$OBHC, tested on OBHC. Bold: best performance.}
    \label{table:DRYAD-OBHC}
\end{table*}


\subsection{Domain Shift}
To quantify domain shift between datasets and evaluate existing adaptation methods, we conducted comprehensive experiments summarised in Tables \ref{domain_shift_MNIST} and \ref{domain_shift_MRI}. \\We first established baseline performance by training and testing segmentation networks within each domain, which served as the oracle performance. These within-domain models achieved average Dice scores exceeding 0.9 for Morpho-MNIST and above 0.8 across all MRI datasets (Tables \ref{domain_shift_MNIST} and \ref{domain_shift_MRI}, row 1).
When evaluating out-of-domain (OoD) performance, we observed substantial degradation in segmentation accuracy across all target domains: Morpho-MNIST thick$\rightarrow$thin yielded DSC=0.669 (Table \ref{domain_shift_MNIST}, row 2), while DRYAD$\rightarrow$HarP achieved DSC=0.705 and DRYAD$\rightarrow$OBHC showed an even more pronounced decline to DSC=0.550 (Table \ref{domain_shift_MRI}, row 2). These results confirm that larger domain gaps correlate with greater performance deterioration.
We subsequently evaluated common image-space augmentation strategies, specifically intensity and geometric augmentations, to address style and content shift, respectively. For Morpho-MNIST, random augmentations failed to improve upon baseline OoD performance. In contrast, for MRI datasets, both intensity augmentations and affine transformations provided marginal improvements over baseline OoD performance for DRYAD$\rightarrow$HarP, with more substantial gains observed for DRYAD$\rightarrow$OBHC (Table \ref{domain_shift_MRI}, rows 3-5 vs. row 2).

Since existing image-space UDA approaches are not explicitly designed to handle content shift, we evaluated a representative feature-based UDA method for segmentation \cite{dinsdale_deep_2021} as a comparative baseline (Tables \ref{domain_shift_MNIST} and \ref{domain_shift_MRI}, row 6). In the case of Morpho-MNIST, the feature-space UDA performed comparably to random augmentations in image space. Alternatively, for the MRI datasets, this approach yielded inferior Dice scores compared to the OoD baseline, with random augmentations demonstrating superior performance. These findings establish a comprehensive benchmark for evaluating our proposed image-space UDA framework, which specifically targets content shift alignment.

\subsection{Content Alignment \& Downstream Segmentation} 
Tables \ref{table:MNIST}, \ref{table:DRYAD-HARP} and \ref{table:DRYAD-OBHC} present quantitative evaluation of our method across content alignment (CA) and downstream segmentation (DS) tasks for three experimental scenarios: Morpho-MNIST thick$\rightarrow$thin, DRYAD$\rightarrow$ HarP and DRYAD$\rightarrow$OBHC, respectively. We evaluate the contribution of each proposed component through ablation studies that incrementally incorporate the discriminator loss  $\mathcal{L}_{disc}$ and segmenter loss $\mathcal{L}_{seg}$ into the training framework. 

Our results demonstrate consistent performance improvements across all datasets when both discriminator and segmenter losses are integrated during training. For Morpho-MNIST (Table \ref{table:MNIST}), our content alignment network improved Dice scores from 0.758 (standard DIR) to 0.818 (row 1 vs. row 4). Correspondingly, we achieved substantial improvements in downstream segmentation performance, with Dice scores increasing from 0.669 (OoD baseline) to 0.830 using our complete framework (row 6 vs. row 10). 

For DRYAD$\rightarrow$ HarP (Table \ref{table:DRYAD-HARP}), our content alignment network enhanced Dice scores from 0.566 (standard DIR) to 0.619 (row 1 vs. row 4), while downstream segmentation performance improved from 0.705 (without CA) to 0.771 (row 6 vs. row 10), surpassing random augmentation-based approaches (0.717) reported in Table \ref{Table:Harp_BHC_domain_shift}.

For MRI, the most substantial improvements were observed in the clinically relevant DRYAD$\rightarrow$OBHC scenario (Table \ref{table:DRYAD-OBHC}), where our approach enhanced content alignment from 0.514 to 0.575 (row 1 vs. row 4) and downstream segmentation from 0.550 to 0.680 (row 6 vs. row 10). This represents a 24\% relative improvement over the baseline without CA and a 15\% relative improvement over the best random augmentation-based methods (0.592), demonstrating the effectiveness of our approach in addressing content shift in realistic clinical settings characterised by significant hippocampal atrophy.

Examples of registration across each dataset are shown in Figure \ref{fig:reg_all}, along with their corresponding segmentation masks and deformation fields. The final column displays the learned deformation fields for DRYAD$\rightarrow$HarP (middle row) and DRYAD$\rightarrow$OBHC (bottom row), where the high-magnitude deformations (bright regions) are concentrated around the hippocampal structures. This spatial distribution of deformation validates our joint training approach, demonstrating that the segmentation-guided registration network effectively prioritises anatomical regions of interest. The concentration of deformation around the hippocampi reflects the network's learned ability to focus computational resources on regions with the greatest anatomical variability, as enforced by the segmentation loss $\mathcal{L}_{seg}$ that backpropagates through the registration network during joint optimisation.
\begin{figure}[b]
\centering
\includegraphics[width=0.95\linewidth]{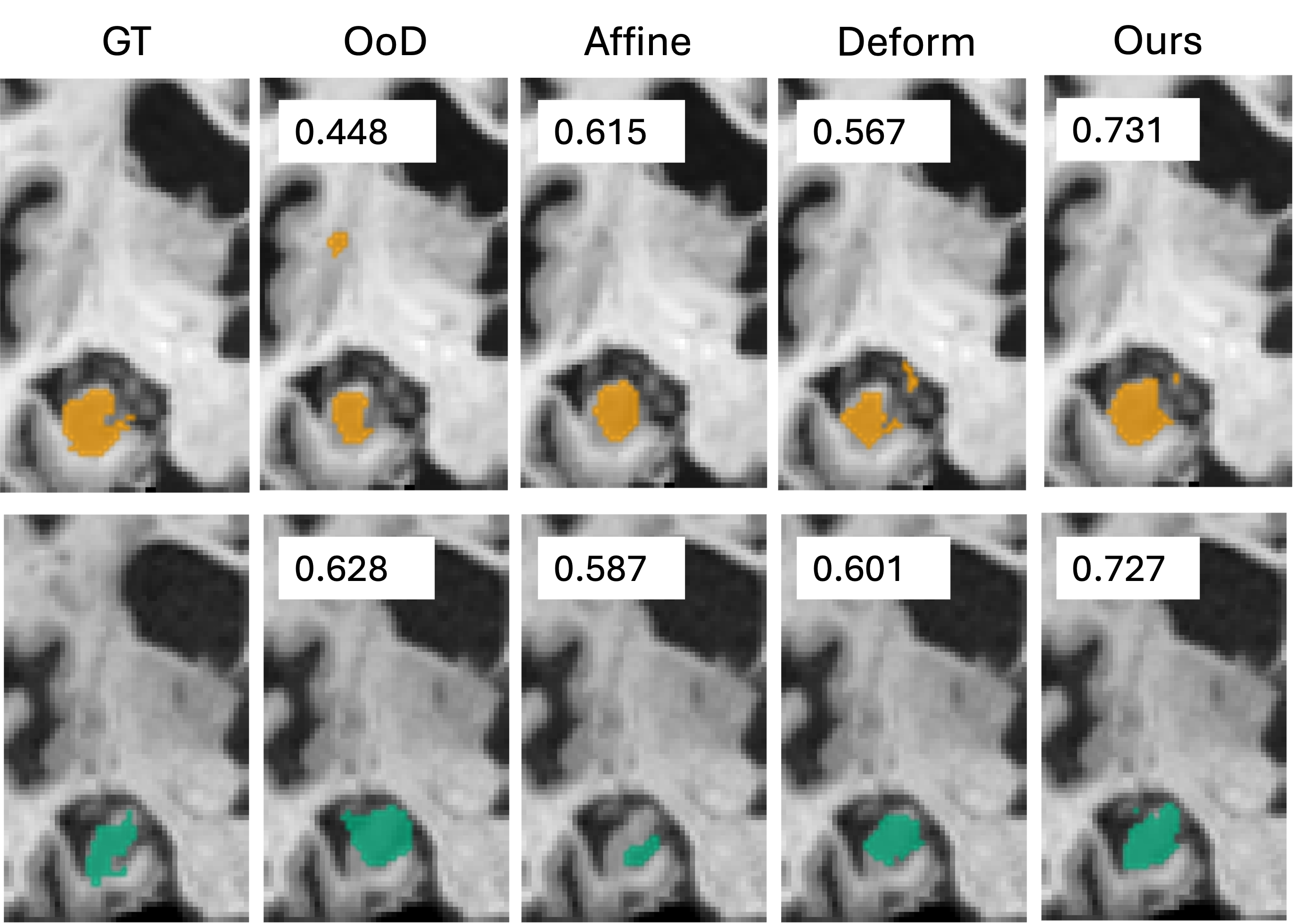}
\caption{Predicted segmentation masks and corresponding dice scores comparing our method to image-space baselines tested on a target sample from HarP (orange) and OBHC (green). GT: Ground Truth; OoD: Out-of-Domain; Affine: Affine augmentations; Deform: Deformable augmentation.}
\label{figure:seg_masks}
\end{figure}
Figure \ref{figure:seg_masks} presents representative segmentation results on target domain samples from HarP (top row) and OBHC (bottom row) datasets. The first column displays the manual ground truth annotations, while subsequent columns show predictions from segmentation models trained under different data augmentation strategies. Specifically, we compare models trained on: (i) source DRYAD data without augmentation (second column: OoD), (ii) DRYAD with standard affine transformations (third column: Affine), (iii) DRYAD with deformable augmentation (fourth column: Deform), and (iv) DRYAD enhanced with target-like images generated by our proposed content alignment (CA) network for DRYAD$\rightarrow$HarP and DRYAD$\rightarrow$OBHC transfers (fifth column: Ours). The results demonstrate that segmentation models trained on content-aligned data achieve superior performance both quantitatively, as evidenced by higher Dice coefficients, and qualitatively, through improved preservation of hippocampal morphology and boundary delineation.
\section{Conclusion}
In this paper, we introduced an image-space UDA framework that tackles both \emph{style} and \emph{content} shifts between source and target domains for improved hippocampus segmentation. We harmonised MRI appearance variations through intensity and spatial preprocessing to address style shift. We also developed a novel content alignment strategy utilising bidirectional deformable image registration with region-specific attention and adversarial constraints to generate target-like transformations. Extensive validation across diverse datasets showed marked performance gains, particularly in the challenging clinical scenario, where our method surpassed augmentation-based approaches by 15\%. These findings confirm our framework's efficacy in overcoming domain adaptation challenges, especially in clinical contexts featuring substantial hippocampal atrophy, thereby advancing automated hippocampus segmentation for real-world applications. 


\section*{Acknowledgements} The authors are grateful for support from: the University of Oxford Department of Computer Science Scholarship (HK), the Bill and Melinda Gates Foundation (NKD, AILN) and the Presidential Postdoctoral Fellowship (Nanyang Technological University) (PHY). We are grateful to the operations team of the OBHC. The OHBC data collection and analysis is supported by the NIHR Oxford Health Biomedical Research Centre (NIHR203316) - a partnership between the University of Oxford and Oxford Health NHS Foundation Trust, the NIHR Oxford Cognitive Health Clinical Research Facility, and the Wellcome Centre for Integrative Neuroimaging (203139/Z/16/Z, 203139/A/16/Z). The views expressed are those of the author(s) and not necessarily those of the NIHR or the Department of Health and Social Care. For the purpose of open access, the authors have applied a CC BY public copyright licence to any Author Accepted Manuscript version arising from this submission.

\bibliographystyle{IEEEtran}
\bibliography{main}

\end{document}